\documentclass[10pt,twocolumn,letterpaper]{article}

\usepackage{cvpr}
\usepackage{times}
\usepackage{epsfig}
\usepackage{graphicx}
\usepackage{amsmath}
\usepackage{amssymb}
\usepackage{subcaption, booktabs}
\usepackage{xcolor}
\usepackage{multirow}
\usepackage{caption}

\graphicspath{{./figures/}}

\usepackage[pagebackref=true,breaklinks=true,letterpaper=true,colorlinks,bookmarks=false]{hyperref}

\cvprfinalcopy 

\def\cvprPaperID{22} 

\ifcvprfinal\fi
\begin{document}

\title{FaR-GAN for One-Shot Face Reenactment}

\author{\parbox{16cm}{\centering
    {\large Hanxiang Hao$^1$, Sriram Baireddy$^1$, Amy R. Reibman$^2$, Edward J. Delp$^1$}\\
    {\normalsize
    $^1$ Video and Image Processing Lab (VIPER),  School of Electrical and Computer Engineering, \\ Purdue University, West Lafayette, Indiana USA\\
    $^2$ School of Electrical and Computer Engineering, Purdue University, West Lafayette, Indiana USA}}
}

\twocolumn[{
\renewcommand\twocolumn[1][]{#1}
\maketitle
\begin{center}
   \centering
	\includegraphics[width=0.9\linewidth]{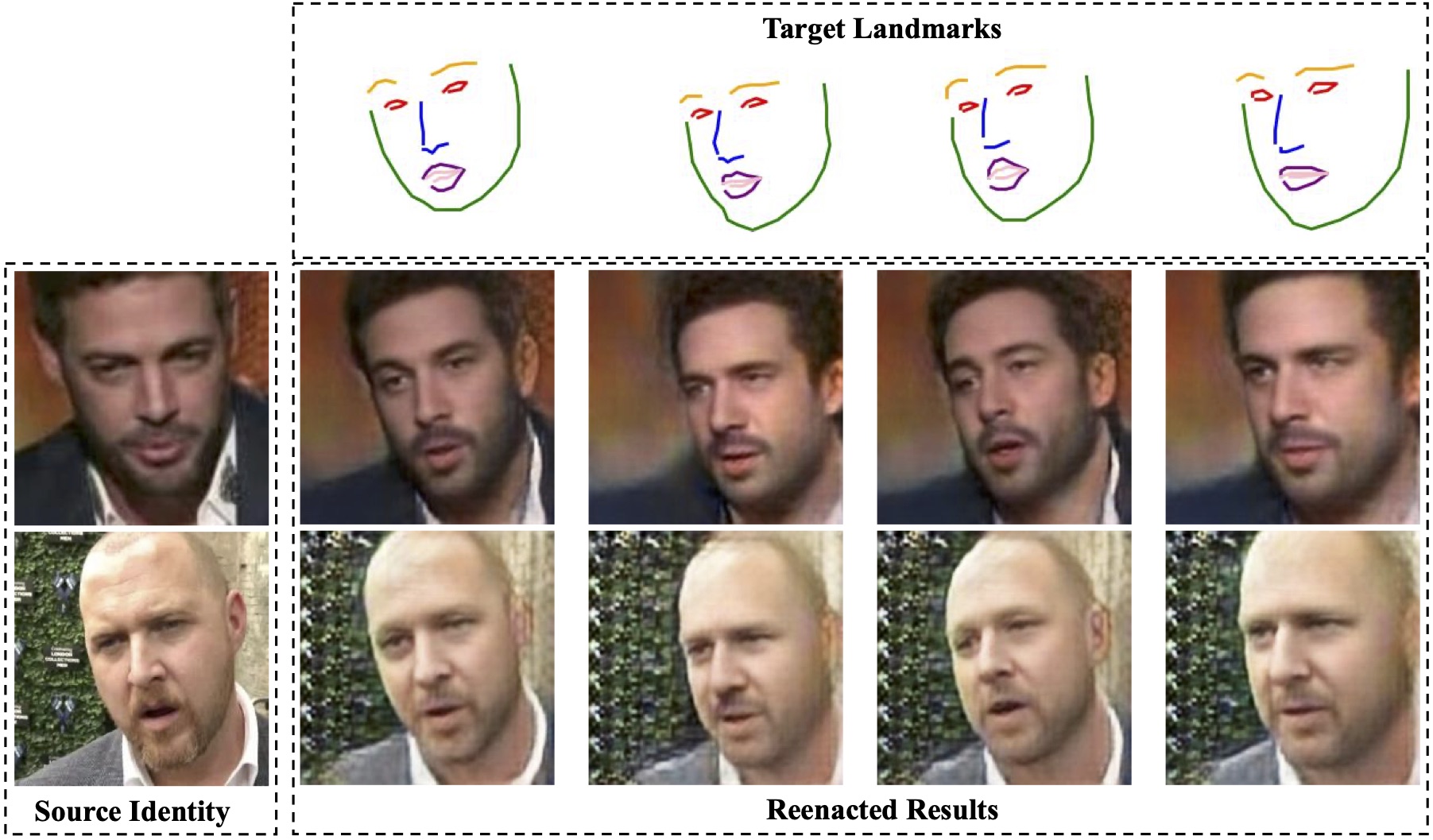}
	\captionof{figure}{\textbf{One-shot face reenactment results from the proposed model.} The proposed method takes a face image from a source identity (visualized on the left colomn) and a target landmark mask (visualized on the top row), and then outputs the face of the source identity but with the target expression.}
	\label{fig:intro}
\end{center}
}]


\begin{abstract}
   Animating a static face image with target facial expressions and movements is important in the area of image editing and movie production.
   This face reenactment process is challenging due to the complex geometry and movement of human faces. 
   Previous work usually requires a large set of images from the same person to model the appearance. 
   In this paper, we present a one-shot face reenactment model, FaR-GAN, that takes only one face image of any given source identity and a target expression as input, and then produces a face image of the same source identity but with the target expression.
   The proposed method makes no assumptions about the source identity, facial expression, head pose, or even image background.
   We evaluate our method on the VoxCeleb1 dataset and show that our method is able to generate a higher quality face image than the compared methods.
\end{abstract}

\section{Introduction} \label{sec:intro}

With the rapid development of generative models in computer vision, especially generative adversarial networks (GANs), there has been an increasing focus on challenging tasks, such as realistic photograph generation, image-to-image translation, text-to-image translation, and super resolution.
Face reenactment is one of these challenging tasks that requires 3D modeling of the geometry and movement of faces.
It has many applications in image editing/enhancement and interactive systems (e.g., animating an on-screen agent with natural human poses/expressions)~\cite{Wiles_2018}.
Producing photo-realistic face reenactment requires a large amount of images from the same identity for appearance modeling.
In this paper, we focus on a more challenging task, one-shot face reenactment, that only requires one image of a given identity to produce a photo-realistic face.
More specifically, we propose a deep learning model that takes one image from a random source identity alongside target expression landmarks.
The model then outputs a face image that has the same appearance information as the source identity, but with the target expression.
This requires the model to transform a source face shape (e.g., facial expression and pose) to a target face shape, while also simultaneously preserving the appearance and the identity of the source face, and even the background.

Figure \ref{fig:intro} shows the reenacted faces produced by the proposed method.
Given an input face image of a source identity, the proposed one-shot face reenactment model, FaR-GAN, is able to transform the expression from the input image to any target expression.
The reenacted faces have the same expression captured by the target landmarks, while also retaining the same identity, background, and even clothes as the input image.
Therefore, the proposed one-shot face reenactment model requires no assumption about the source identity, facial expression, head pose, and image background.

The main contributions of this paper are summarized as follows:
\begin{itemize}
   \item We develop a GAN-based method that addresses the task of one-shot face reenactment. 
   \item The proposed FaR-GAN is able to compose appearance and expression information for effective face modeling.
   \item The reenacted images produced by the proposed method achieve higher image quality than the compared methods.
\end{itemize}

\section{Related Work} \label{sec:related_work}

\textbf{Face Reenactment by 3D Modeling.}
Modeling faces in 3D helps in accurately capturing their geometry and movement, which in turn improves the photorealism of any reenacted faces.
Thies \etal~\cite{Thies_2018} propose a real-time face reenactment approach based on the 3D morphable face model (3DMM)~\cite{Blanz_1999} of the source and target faces.
The transfer is done by fitting a 3DMM to both faces and then applying the expression components of one face onto the other~\cite{Nirkin_2019}.
To achieve face synthesis based on imperfect 3D model information, they further improve their method by introducing a learnable feature map (\ie neural texture) alongside the UV map from the coarse 3D model as input to the rendering system~\cite{Thies_2019}. 
During 2D rendering, they also design a learnable neural rendering system that is based on U-Net~\cite{Ronneberger_2015} to output the 2D reenacted image.
The entire rendering pipeline is end-to-end trainable.

\textbf{Face Reenactment by GANs.}
Generative adversarial networks have been successfully used in this area due to their ability to generate photo-realistic images.
They are able to achieve high quality and high resolution unconditional face generation~\cite{Karras_2018, Karras_2019,Karras_2019b}.
ReenactGAN, proposed by Wu \etal~\cite{Wu_2018}, first maps the face that contains the target expression into an intermediate boundary latent space that contains the information of facial expressions but no identity-related information.
Then the boundary information is used for an identity-specific decoder network to produce the reenacted face of the specific identity.
Therefore, their model cannot be used for the reenactment of unknown identities.

To solve this issue, few-shot or even one-shot face reenactment methods have also been developed in the recent work~\cite{Wiles_2018, Zakharov_2019, Zhang_2019b}.
Wiles \etal~\cite{Wiles_2018} propose a model, namely X2Face, that is able to use facial landmarks or audio to drive the input source image to a target expression.
Instead of directly learning the transformation of expressions, their model first learns the frontalization of the source identity. 
“Frontalization” is the process of synthesizing frontal facing views of faces appearing in single unconstrained photos~\cite{Hassner_2015}.
Then it produces an intermediate interpolation map given the target expression to be used for transferring the frontalized face.
Zakharov \etal~\cite{Zakharov_2019} present a few-shot learning approach that achieves the face reenactment given a few, or even one, source images.
Unlike the X2Face model, their method is able to directly transfer the expression without the intermediate boundary latent space~\cite{Wu_2018} or interpolation map~\cite{Wiles_2018}.
Zhang \etal~\cite{Zhang_2019b} propose a one-shot face reenactment model that only requires one source image for training and inferencing.
They use an auto-encoder-based structure to learn the latent representation of faces, and then inject these features using the SPADE module~\cite{Park_2019} for the face reenactment task.
The SPADE module in our proposed method is inspired by their work. 
However, instead of using the multi-scale landmark masks used by~\cite{Zhang_2019b}, we use learnable features from convolution layers as the input to the SPADE module.

\section{Proposed Method} \label{sec:proposed_method}

\subsection{Model Architecture}

\begin{figure*}[t]
  \begin{center}
    \includegraphics[width=0.9\linewidth]{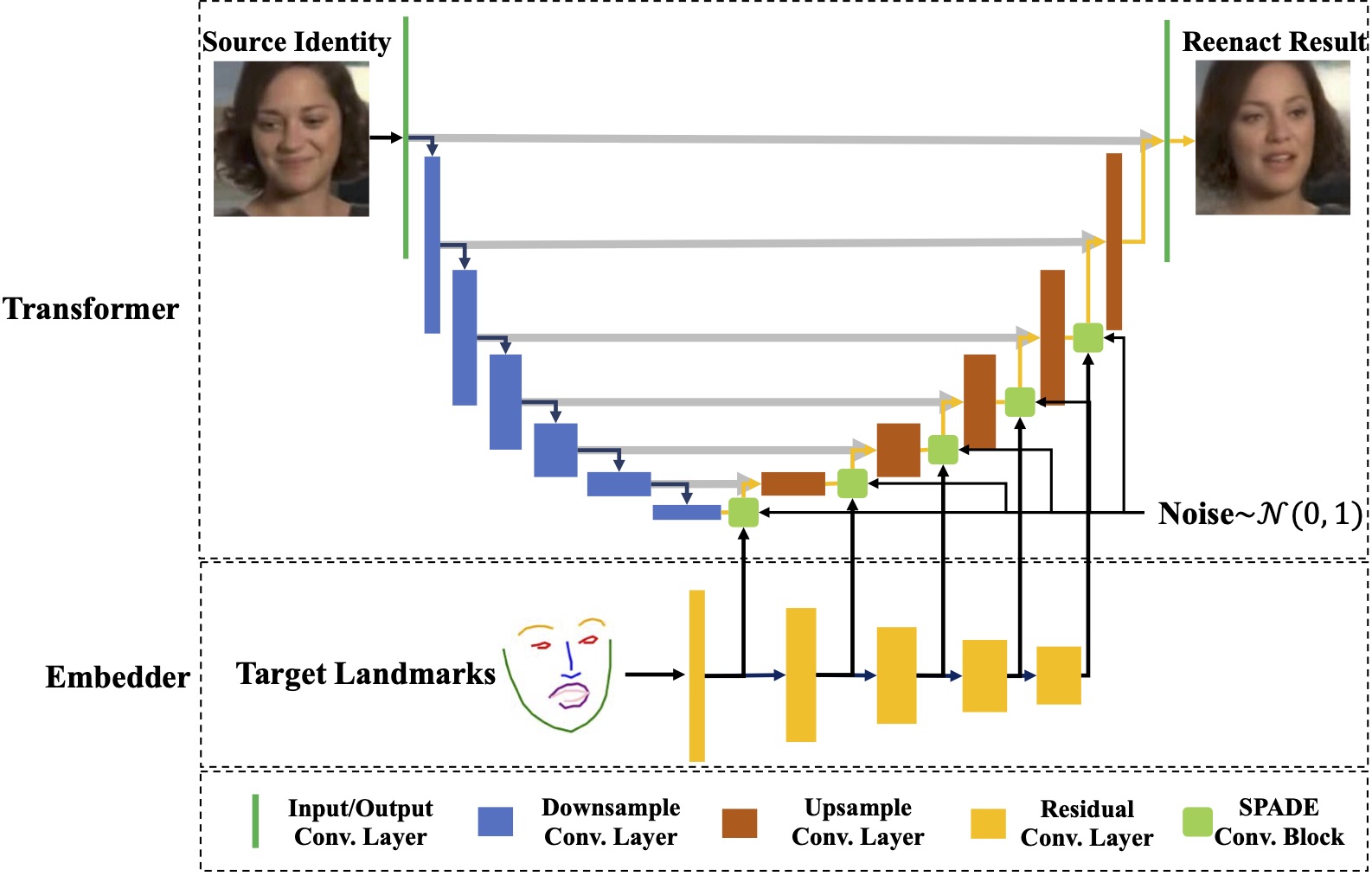}  
  \end{center}
    \caption{\textbf{The generator architecture of the proposed FaR-GAN model.} Given target facial landmarks and a arbitrary source identity, the proposed model learns to transfer facial expression for the source identity. The embedder model learns the feature representation of the facial expression defined by the landmarks. The transformer model uses the features from the embedder to generate a new face of the source identity but has the same facial expression as the target landmarks.}
  \label{fig:architecture}
\end{figure*}

Figure \ref{fig:architecture} shows the generator architecture of the proposed FaR-GAN model. 
The model consists of two parts: embedder and transformer.
The embedder model aims to learn the feature representation of facial expressions given a set of facial landmarks.
In this work, we adopt a similar color encoding method proposed in~\cite{Zakharov_2019} to represent the facial landmarks.
More specifically, we use distinct colors for eyes, eyebrows, nose, mouth outlier, mouth inlier, and face contour.
We also tried to use a binary mask to represent the landmark information (\ie set 1 for the facial region and set 0 for the background), but it did not give us a better result.
We will show the comparison of results with different landmark representations in Section \ref{sec:experiments}.
The transformer model aims to use the landmark features from the embedder model to reenact the input source identity with the target landmarks. 
The transformer architecture is based on the U-Net model~\cite{Ronneberger_2015}.
The U-Net model is a fully convolutional network for image segmentation. 
Besides its encoder-decoder structure for local information extraction, it also utilizes skip connections (the gray arrows in Figure \ref{fig:architecture}) to retain global information.

A similar generator architecture can be found in~\cite{Zakharov_2019} but with several differences.
First, instead of using the embedder to encode appearance information of the source identity, we use it to extract the target landmark information.
The embedder model is a fully convolutional network that continuously downsamples the feature resolution with maxpooling or average-pooling layers.
Therefore, the spatial information of the input image will be lost due to the downsampling process.
To encode the appearance information of the source identity, the output features are required to represent a large amount of information including the identifiable information, hair style, body parts (neck and shoulders), and even background.
Therefore, it is challenging for the embedder model to learn precise appearance information with the loss of the spatial information.
In our approach, we use the embedder model to encode the facial landmarks, which contains much less information than the aforementioned appearance features.
Moreover, instead of outputting a single 1D embedding vector~\cite{Zakharov_2019}, we use the embedder features from all resolutions obtained after the downsampling process. 
With this, we can assure the embedder features contain the required spatial information for expression transformation.

\begin{figure}[htbp]
   \begin{center}
     \includegraphics[width=0.8\linewidth]{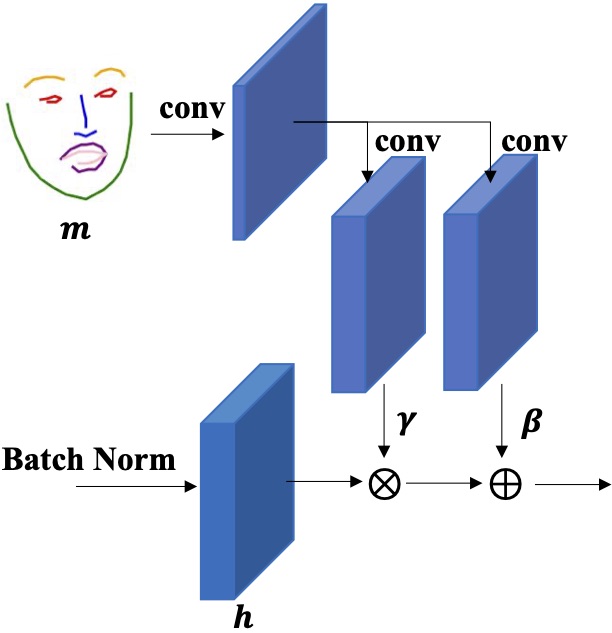}  
   \end{center}
     \caption{\textbf{Architecture of the SPADE module.} This figure is based on~\cite{Park_2019}. SPADE module maps the input landmark mask to the modulation parameters $\gamma$ and $\beta$ through a set of convolution layers. Then the element-wise multiplication and addition are used for $\gamma$ and $\beta$, respectively to the batch-normalized input feature.}
   \label{fig:spade}
 \end{figure}

The adaptive instance normalization (AdaIN) module has been successfully used for face generation in previous work~\cite{Karras_2019, Karras_2019b, Zakharov_2019}.
In~\cite{Zakharov_2019}, they use AdaIN modules to inject the appearance information into the generator model to produce the reenacted face by assigning a new bias and scale of the convolution features based on the embedder features.
However, since we need to inject landmark information, which comes from a sparse landmark mask, we cannot simply adopt the AdaIN module in our method.
This is because, the instance normalization (\eg AdaIN) tends to wash away semantic information when applied to uniform or flat segmentation masks~\cite{Park_2019}, such as our input landmark masks.
Instead, we propose using the spatially-adaptive normalization (SPADE)~\cite{Park_2019} module to inject the landmark information. 
As the name indicates, the SPADE module is a feature normalization approach that uses the learnable spatial information from the input features.
Similar to batch normalization~\cite{Ioffe_2015}, the input convolution features are first normalized in a channel-wise manner, and then modulated with a learned scale and bias, as shown in Figure \ref{fig:spade}.
The output of the SPADE module can be formulated, as shown in Equation \ref{eq:spade}.
\begin{equation}\label{eq:spade}
   \gamma_{c,x,y}(\textbf{m}) \frac{h_{n,c,x,y} - \mu_{c}}{\sigma_{c}} + \beta_{c,x,y}(\textbf{m})
\end{equation}
where $\textbf{m}$ is the input landmark mask or intermediate convolution features from the embedder, $h_{n,c,x,y}$ is the input convolution feature from mini-batch $n \in N$, channel $c \in C$, dimension $x \in W$, and dimension $y \in H$, $\gamma_{c,x,y}$ is the new scale, and $\beta_{c,x,y}$ is the new bias.
The mean $\mu_{c}$ and standard deviation $\sigma_{c}$ of the activation in channel $c$ are defined in Equation \ref{eq:mean} and \ref{eq:std}.
\begin{equation}\label{eq:mean}
   \mu_{c} = \frac{1}{NHW}\sum_{n,x,y}h_{n,c,x,y}
\end{equation}
\begin{equation}\label{eq:std}
   \sigma_{c} = \sqrt{\frac{1}{NHW}\sum_{n,x,y}(h_{n,c,x,y}^2 - \mu_{c}^2})
\end{equation}

\begin{figure}[htbp]
   \begin{center}
     \includegraphics[width=1\linewidth]{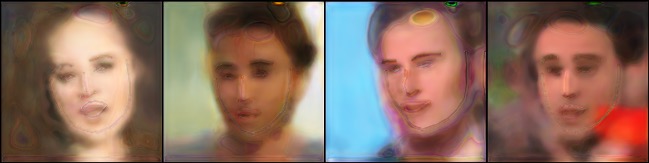}  
   \end{center}
     \caption{\textbf{The artifacts of using multi-scale masks as input to the SPADE module.} The landmark contours are still visible in the output images of the transformer model.}
   \label{fig:spade_artifact}
 \end{figure}

This SPADE module has been successfully used for the face reenactment task in~\cite{Zhang_2019b}.
As shown in Figure \ref{fig:architecture}, in our method, the input to the SPADE block is the convolution features from the embedder network.
In~\cite{Zhang_2019b}, they use a group of multi-scale landmark masks as the input to the SPADE blocks, instead of the deep features from our proposed method.
However, in our experiment, if we use these multi-scale masks instead of deep features as input to the SPADE blocks, the output reenacted faces will contain the artifacts from the input landmark contours, as shown in Figure \ref{fig:spade_artifact}.
Similar to~\cite{Liu_2019}, we use the features from the embedder network to inject the landmark information into the transformer model.

There are many aspects in human portraits that can be regarded as stochastic, such as the exact placement of hairs, stubble, freckles, or skin pores~\cite{Karras_2019}.
Inspired by StyleGAN~\cite{Karras_2019}, we introduce stochastic variation into our transformer model by injecting noise.
The noise injection is executed for each resolution of the decoder part of the transformer model.
More specifically, we first sample an independent and identically distributed standard Gaussian noise map $z$ of size $H \times W$, where $H$ and $W$ are the spatial resolution of the input feature.
Then a noise block with the number of channels $C$ is obtained by scaling the noise map $z$ with a set of learnable scaling factors for each channel.
We inject the noise block by adding it element-wise with the input features.

We adopt the design of~\cite{Mirza_2014, Isola_2016} for our discriminator.
More specifically, the input to our discriminator is the reenacted face concatenated with the target landmark mask, or the ground truth face image with its corresponding landmark mask.
Therefore, the discriminator aims to guide the generator to produce a realistic face and also faces with the correct target landmarks.
In Section \ref{sec:experiments}, we will provide an ablation study to show the importance of the discriminator.

\subsection{Loss Function}

The proposed model including both embedder and transformer is trained end-to-end.
Assume we have a set of videos that contain the moving face/head of multiple identities.
We denote  $\textbf{x}_i(t)$ as the $i$-th video and $t$-th frame.
Assume $\textbf{x}_i(t_1)$ and $\textbf{x}_i(t_2)$ are two random frames from a video.
Therefore, the two frames $\textbf{x}_i(t_1)$ and $\textbf{x}_i(t_2)$ contain the same identity but with different facial expressions and head poses.
We formulate our generator function $G$ as follows:
\begin{equation}
   \hat{\textbf{x}}_i(t_2) = G(\textbf{x}_i(t_1), \textbf{m}_i(t_2)) 
\end{equation}
where $\textbf{m}$ is the landmark mask.
The generator loss function is defined in Equation \ref{eq:g_loss}.
\begin{equation}\label{eq:g_loss}
   \mathcal{L}_G = \mathcal{L}_{adv} + \mathcal{L}_{L1} + \mathcal{L}_{p} + \mathcal{L}_{id}
\end{equation}
where
\begin{align*}
   &\mathcal{L}_{adv} = \mathbb{E}_{\textbf{x}_i, \textbf{m}_i}[(D(\hat{\textbf{x}}_i(t_2), \textbf{m}_i(t_2)) - 1)^2]\\
   &\mathcal{L}_{L1}     = \Vert \hat{\textbf{x}}_i(t_2) - \textbf{x}_i(t_2) \Vert \\
   &\mathcal{L}_{p}      = \sum_{l \in \Phi} \Vert \phi_l(\hat{\textbf{x}}_i(t_2)) - \phi_l(\textbf{x}_i(t_2)) \Vert \\
   &\mathcal{L}_{id}     = \sum_{l \in \Psi} \Vert \psi_l(\hat{\textbf{x}}_i(t_2)) - \psi_l(\textbf{x}_i(t_2)) \Vert.
\end{align*}
$\mathcal{L}_{adv}$ is the generator adversarial loss, which is based on LSGAN~\cite{Mao_2017}.
We compared the results from the vanilla-GAN~\cite{Goodfellow_2014}, LSGAN~\cite{Mao_2017}, and WGAN-GP~\cite{Gulrajani_2017} and chose LSGAN based on the visual quality of reenacted images.
$\mathcal{L}_{L1}$ is the pixel-wise L1 loss to minimize the pixel difference of the generated image and the ground truth image.
$\mathcal{L}_{p}$ is the perceptual loss for minimizing the semantic difference, which was originally proposed by~\cite{Johnson_2016}.
$\Phi$ is a collection of convolution layers from the perceptual network and $\phi_l$ is the activation from the $l$-th layer. 
In our work, the perceptual network is a VGG-19 model~\cite{Simonyan_2015} pretrained on the ImageNet dataset~\cite{Russakovsky_2015}.
To enforce the reenacted face to have the same identifiable information as the input source identity, we add an identity loss $\mathcal{L}_{id}$, which is similar to the perceptual loss, but with a VGGFace model~\cite{Cao_2018} pretrained for face verification.

The discriminator loss function is based on the LSGAN loss function, which is defined as follows:
\begin{equation}\label{eq:d_loss}
   \begin{aligned}
      \mathcal{L}_D = &\mathbb{E}_{\textbf{x}_i, \textbf{m}_i}[(D(\hat{\textbf{x}}_i(t_2), \textbf{m}_i(t_2)))^2] + \\
                      &\;\;\;\; \mathbb{E}_{\textbf{x}_i, \textbf{m}_i}[(D(\textbf{x}_i(t_2), \textbf{m}_i(t_2)) - 1)^2]
   \end{aligned}
\end{equation}

\subsection{Implementation Details}

As shown in Figure \ref{fig:architecture}, the convolution layers in the embedder model are a set of residual convolution layers~\cite{He_2016}.
This is adopted from~\cite{Zakharov_2019}, which also adds the spectral normalization~\cite{Miyato_2018} layers to stabilize the training process.
The transformer network consists of input/output convolution layers, downsampling convolution layers, upsampling convolution layers, and SPADE convolution blocks.
The input/output convolution layers only contain convolution layers; so the feature resolutions do not change. 
The downsampling convolution layers consist of an average-pooling layer, convolution layer, and spectral normalization layer. 
The upsampling convolution layers consist of a de-convolution layer followed by a spectral normalization layer to upsample the feature resolution by a factor of 2. 
The SPADE convolution block contains the noise injection layer followed by the SPADE module. 
For the discriminator, we use the same structure proposed by~\cite{Isola_2016}, with the two downsampling convolution layers.

Previously, the self-attention mechanism has been successfully used for GANs that generate high quality synthetic images~\cite{Zhang_2019}.
To ensure that the generator learns from a long-range of information within the entire input image, we adopt the self-attention module in both the generator and discriminator.
More specifically, for the generator, we place the self-attention module after the upsampling convolution layers of the feature resolutions of $32 \times 32$ and $64 \times 64$, which is similar to the implementation in~\cite{Zakharov_2019}.
For the discriminator, we place the self-attention module after the second downsampling convolution layer.

During training, in order to balance the magnitude of each term in the loss function, we choose the weights for $\mathcal{L}_{L1}$, $\mathcal{L}_{p}$, and $\mathcal{L}_{id}$ as 20, 2, and 0.2, respectively.
These weights could be different when using different datasets or different perceptual networks.
We use the Adam optimizer~\cite{Kingma_2015} for both the generator and discriminator with the initial learning rate as $5e^{-5}$.
The learning rate decays linearly and decreases to 0 after 100 epochs.

\section{Experiments} \label{sec:experiments}

\subsection{Dataset}

In this paper, we use the VoxCeleb1 dataset~\cite{Nagrani_2017} for training and testing.
It contains 24,997 videos from 1251 different identities.
The dataset provides cropped face images extracted at 1 frame per second and we resize these images to $256 \times 256$.
\textit{Dlib} package~\cite{King_2009} is used for extracting 68-point facial landmarks.
We split the identities into training and testing sets with the ratio of $8:2$ in order to assure that our model is generalizable to new identities.

\subsection{Experimental Results}

We compare the proposed method against two methods, the X2Face model~\cite{Wiles_2018} and the few-shot talking face generation model (Few-Shot)~\cite{Zakharov_2019}.
X2Face contains two parts: an embedder network and a driver network.
Instead of directly mapping the input source image to the reenacted image, their embedder learns to frontalize the input source image and the driver network produces a interpolation map given the target expression to transform the frontalized image.
To compare with the X2Face model, we use their model with pretrained weights provided by the authors and evaluate on the VoxCeleb1 dataset.
The Few-Shot model also contains two parts: an embedder network and a generator network.
As described in Section \ref{sec:proposed_method}, their embedder learns to encode the appearance information of the source image, while the generator learns to generate the reenacted image given the appearance information and target landmark mask.
For the Few-Shot model, since the authors only provide the testing results, we directly use these results for comparison.
Both the X2Face and Few-Shot method require two stages of training. 
The first stage uses two frames from the same video, while the second stage requires the frames from two different videos. 
By doing so, they can ease the training process at the beginning by using the frames that contain the same identity and similar background information.
Then for the second stage, they use the frames from two different videos to ensure that the reenacted face contains the same identifiable information as the input source identity.
As mentioned in Section \ref{sec:proposed_method}, the proposed method requires only the first stage training.

In this section, we provide both a qualitative and quantitative results comparison.
For the quantitative analysis, we use structured similarity index (SSIM)~\cite{Wang_2004} and Fr\'echet-inception distance (FID)~\cite{Heusel_2017} to measure the quality of the generated images. 
SSIM measures low-level similarity between the ground truth images and reenacted images~\cite{Zakharov_2019}.
The higher the SSIM is, the better the quality of the generated images are.
FID measures perceptual realism based on an InceptionV3 network that was pretrained on ImageNet dataset for image classification (the weights are fixed during the FID evaluation).
It has been used for image quality evaluation in many works~\cite{Karras_2018, Karras_2019}.
In our work, the FID score is computed using the default setting \footnote{The implementation is in https://github.com/mseitzer/pytorch-fid}, so we use the final average pooling features from the InceptionV3 network.
The lower the FID is, the better the quality of the generated images are.

\begin{figure*}[htbp]
   \begin{center}
     \includegraphics[width=0.9\linewidth]{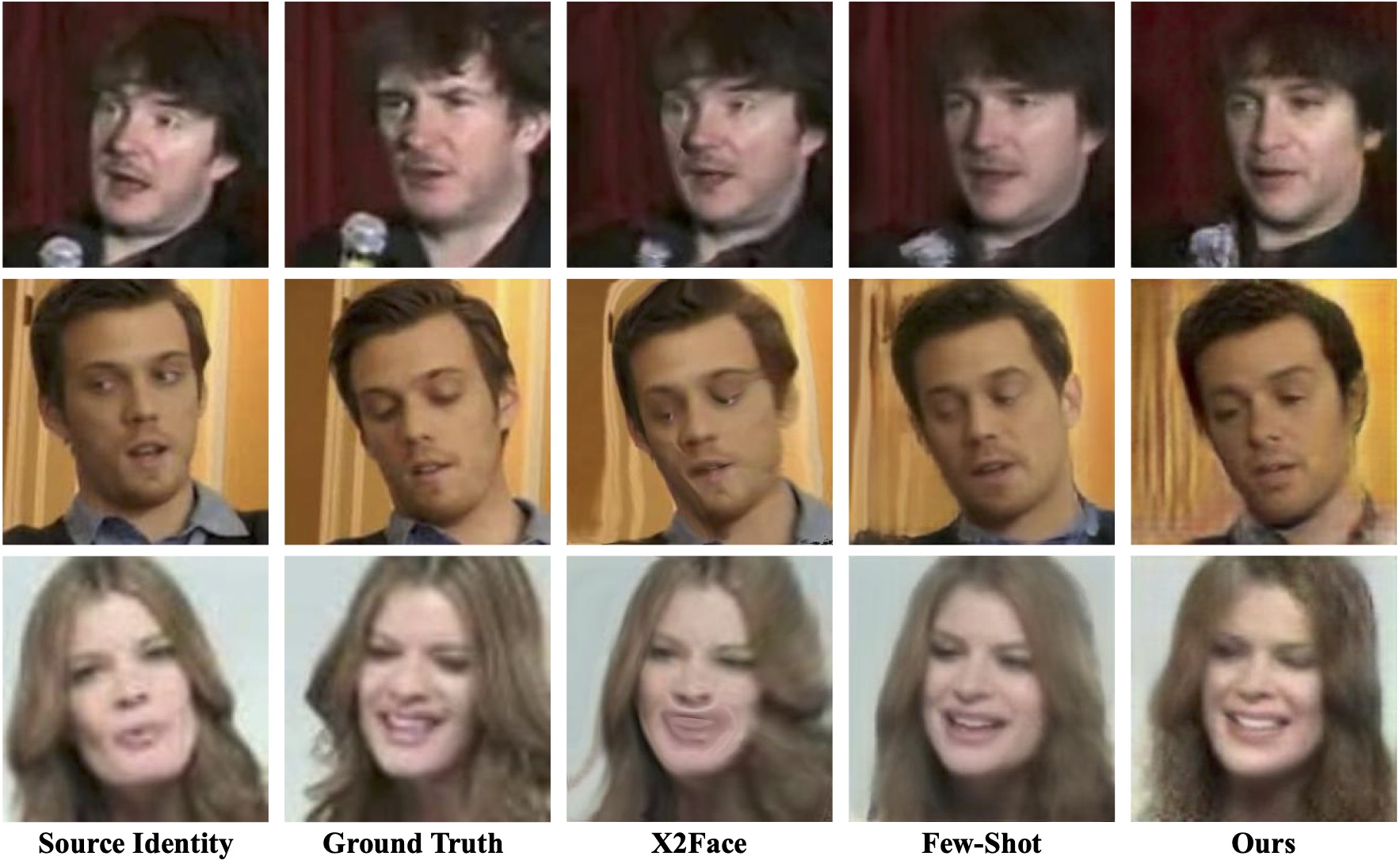}  
   \end{center}
     \caption{\textbf{Face reenactment results from the compared and proposed methods.}}
   \label{fig:compare}
 \end{figure*}

\begin{table}[htbp]
	\begin{center}
		\begin{tabular}{@{}ccc@{}}
			\toprule
			\textbf{Method}            & \textbf{SSIM}$\uparrow$ & \textbf{FID}$\downarrow$ \\ \midrule
			X2Face \cite{Wiles_2018}       & 0.68 & 45.8  \\
			Few-Shot \cite{Zakharov_2019}   & 0.67 & 43.0  \\
			FaR-GAN (proposed)          & 0.68 & 27.1  \\
			\bottomrule
		\end{tabular}
		\caption{\textbf{SSIM and FID results of the compared and proposed methods}.}
		\label{table:results}	
	\end{center}
\end{table}

Table \ref{table:results} shows the results of the proposed and compared methods.
The SSIM and FID scores of the compared methods are obtained from the original paper~\cite{Zakharov_2019}.
Although the SSIM results are similar for all three methods, the proposed method outperforms the compared methods in terms of FID. 
Figure \ref{fig:compare} shows the qualitative comparison from the testing set.
The results from X2Face contains wrinkle artifacts, because it uses the interpolation mask to transfer the source image, instead of directly learning the mapping function from the source image to the reenacted image.
Although the X2Face result in the first row shows its effectiveness when the change of head pose is relatively small, the results in the second and third rows show that the wrinkle artifacts get more visible when the background becomes complex and the change of head pose is larger.
Both Few-Shot method and the proposed method obtain the results with a good visual quality, including transferring accurate target expression and also preserving the background information.
Due to the proposed method of injecting the noise into the transformer network, the reenacted faces contain more high frequency information than the Few-Shot model, especially for the woman's hair from the third testing case.
Because the FID computes the statistical difference from a collection of synthetic images and real images, it measures both high frequency and low frequency components.
Therefore, the proposed method achieves much lower FID than the two compared methods.

\subsection{Ablation Study}

\begin{figure}[htbp]
   \begin{center}
     \includegraphics[width=1\linewidth]{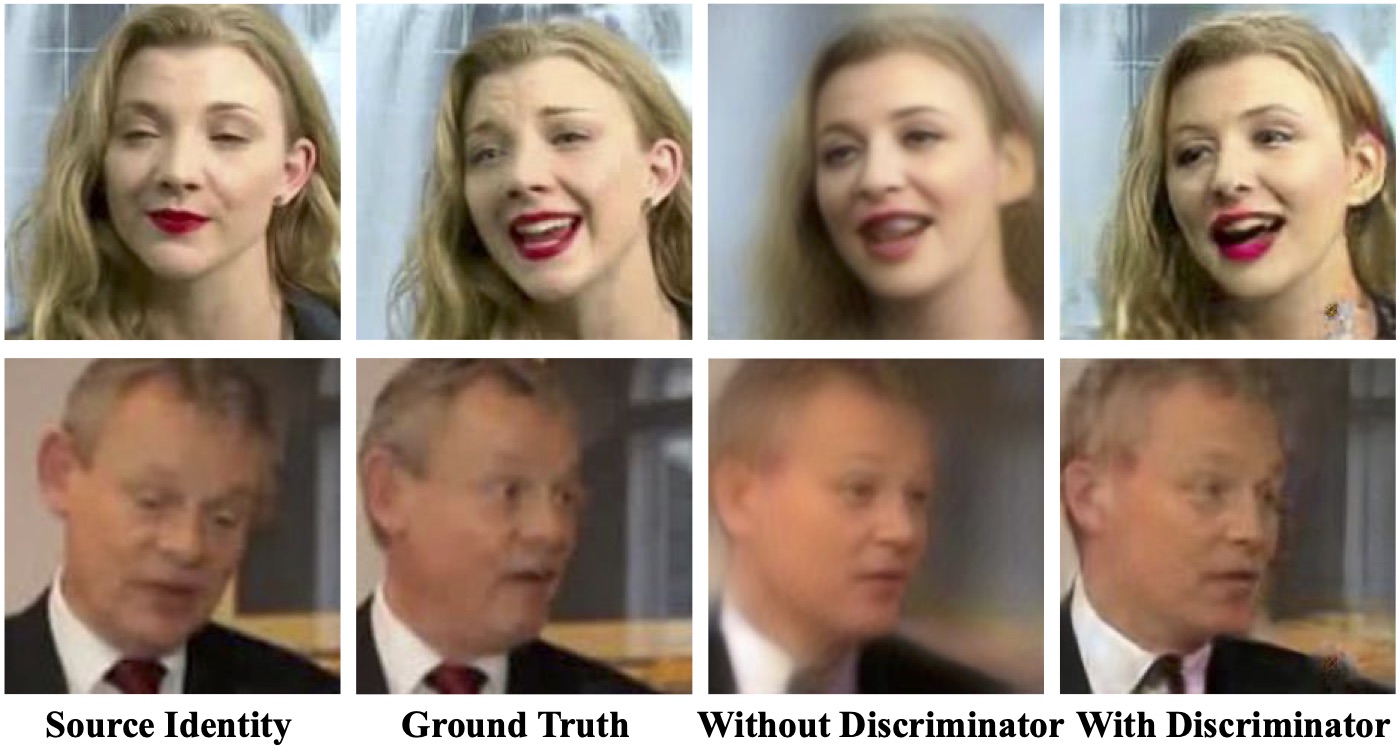}  
   \end{center}
     \caption{\textbf{Ablation study of the use of discriminator.}}
   \label{fig:ablation_discriminator}
 \end{figure}

Figure \ref{fig:ablation_discriminator} shows our results with and without the discriminator.
The result with discriminator contains more details, like hair, teeth, and background, compared to the result without discriminator.
Therefore, the discriminator does guide the generator (both embedder and transformer) in producing better synthetic images.

\begin{table}[htbp]
	\begin{center}
		\begin{tabular}{@{}ccc@{}}
			\toprule
			\textbf{Method}  & \textbf{SSIM}$\uparrow$ & \textbf{FID}$\downarrow$   \\ \midrule
			FaR-GAN (Mask)    & 0.67 & 52.1 \\
			FaR-GAN (Contour) & 0.68 & 27.1 \\
			\bottomrule
		\end{tabular}
		\caption{\textbf{SSIM and FID results of the proposed method with different landmark representations}.}
		\label{table:ablation_landmark}	
	\end{center}
\end{table}

\begin{figure}[htbp]
   \begin{center}
     \includegraphics[width=0.8\linewidth]{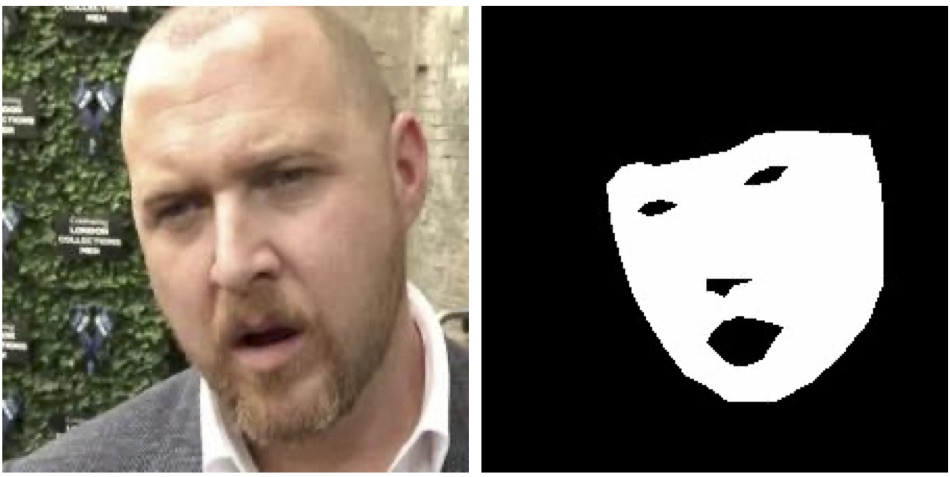}  
   \end{center}
     \caption{\textbf{An alternative landmark representation using a binary mask.}}
   \label{fig:binary_mask}
 \end{figure}

Table \ref{table:ablation_landmark} shows the SSIM and FID results of the model using the contour-based mask and binary mask for landmark representation. 
An example of the landmark binary mask is shown in Figure \ref{fig:binary_mask}.
Although the SSIM scores are similar, the FID score of the binary mask is much higher than the contour-based representation.
Due to the use of different colors for different parts of the facial components, the contour-based mask provides additional information for the embedder to treat different parts of face separately. 
Thus, it can achieve a better understanding of facial pose and expression.

\begin{table}[htbp]
	\begin{center}
		\begin{tabular}{@{}ccc@{}}
			\toprule
			\textbf{Method}                      & \textbf{SSIM}$\uparrow$ & \textbf{FID}$\downarrow$   \\ \midrule
			FaR-GAN (w/o attention and w/o noise) & 0.67 & 63.9 \\
			FaR-GAN (w/ attention and w/o noise)  & 0.66 & 35.3 \\
			FaR-GAN (w/ attention and w/ noise)   & 0.68 & 27.1 \\
			\bottomrule
		\end{tabular}
		\caption{\textbf{SSIM and FID results of the proposed method with different model components}.}
		\label{table:ablation_components}	
	\end{center}
\end{table}

\begin{figure*}[htbp]
   \begin{center}
     \includegraphics[width=1\linewidth]{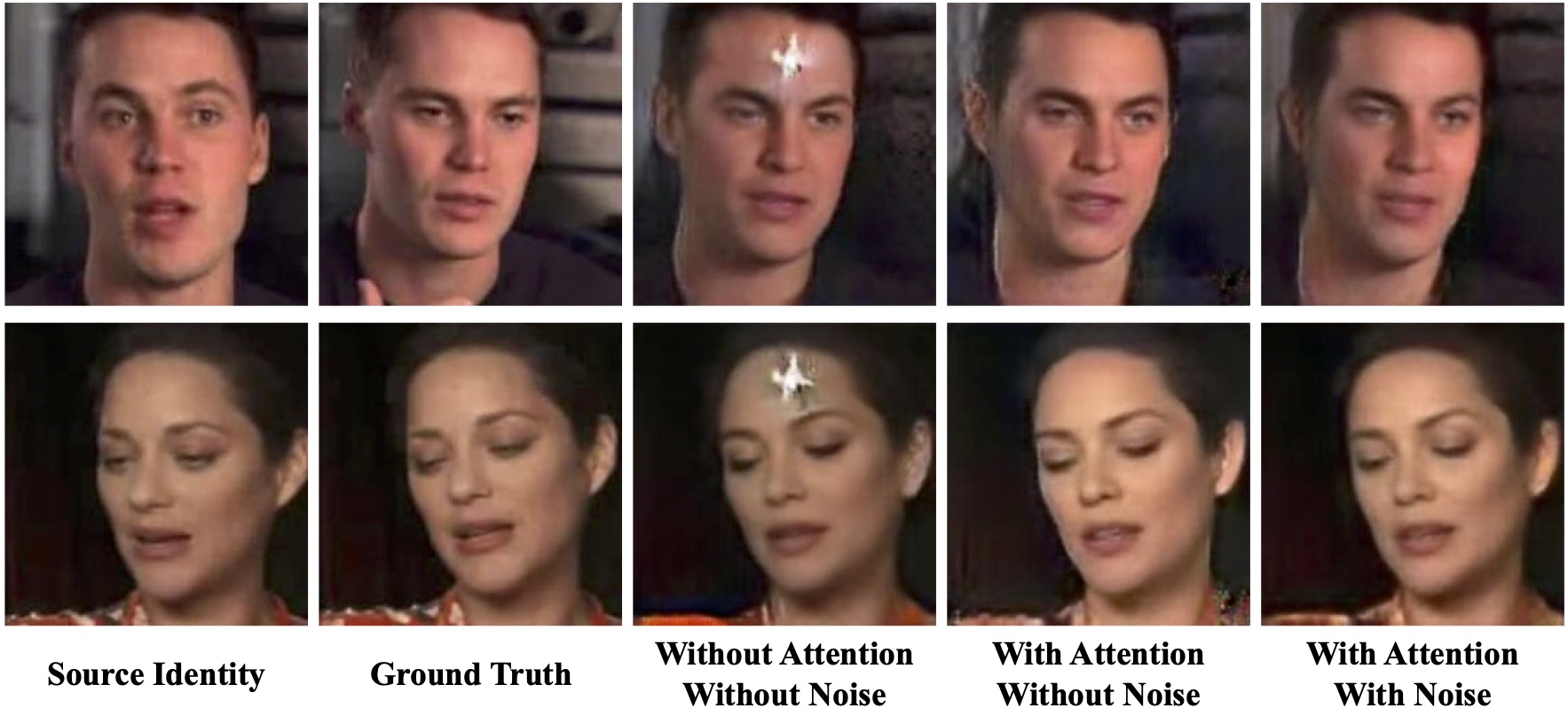}  
   \end{center}
     \caption{\textbf{Ablation study of different model component settings.}}
   \label{fig:ablation_components}
 \end{figure*}

\begin{figure}[htbp]
  \begin{center}
    \includegraphics[width=1\linewidth]{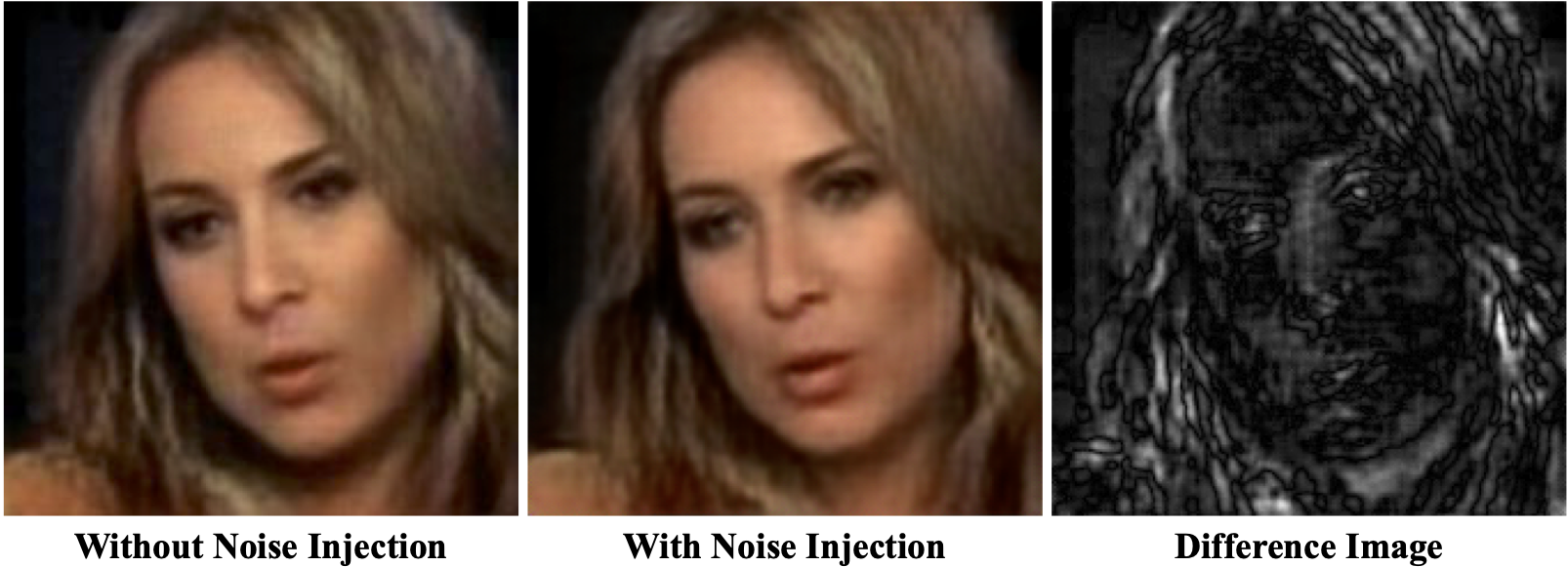}  
  \end{center}
    \caption{\textbf{Noise injection improves the image quality by adding high frequency details.} The difference image has been rescaled for better visualization.}
  \label{fig:noise_injection}
\end{figure}

Table \ref{table:ablation_components} shows the ablation study of different model components, including self-attention and noise injection.
Although the SSIM scores show the similar performance of the three experiments, the FID scores indicate the improvement when using these components.
Adding the self-attention module reduces the FID from 63.9 to 35.3 and with the noise injection module, the FID drops to 27.1. 
Therefore, the two components indeed help improve the model performance.
We also show the visual comparison of these experiments in Figure \ref{fig:ablation_components}.
The results without self-attention and noise injection contain blob-like artifacts that are also mentioned in~\cite{Karras_2019b}.
In general, both of the results with and without noise injection achieve a good visual quality.
However, the results without noise injection have some artifacts, as seen in the ear region in the first example and right shoulder region in the second example.
As shown in Figure \ref{fig:noise_injection}, noise injection can improve the reenacted image quality by adding high frequency details in the hair region.
Therefore, the model with both self-attention and noise injection modules achieves the best image quality.

\section{Conclusion and Future Work} \label{sec:conclusion}

In this paper, we propose a one-shot face reenactment model, FaR-GAN, that is able to transform a face image to the target expression.
The proposed method takes only one image from any identity as well as target facial landmarks and it is able to produce a high quality reenacted face image of the same identity but with the target expression.
Therefore, it makes no assumption about the identity, facial pose, and expression of the input face image and target landmarks.
We evaluate our method using the VoxCeleb1 dataset and show that the proposed model is able to generate face images with better visual quality than the compared methods.

Although the results from our method achieve a high visual quality, in some cases, when the identity that provides the target landmarks has a large appearance difference from the source identity, such as different genders or face sizes, there is still a visible identity gap between the input source identity and the reenacted face.
In future work, we will continue improving our model to bridge this identity gap, such as using an additional finetuning step to explicitly direct the model to reduce the identity changes, as proposed from~\cite{Wiles_2018, Zakharov_2019}.
Furthermore, in the current model setting, we do not consider the pupil movement in our landmark representation. 
As proposed by~\cite{Zhang_2019b}, we can add the gaze information in the landmark mask to make the reenacted face contain more realistic facial movement.
Although the proposed method achieves a good performance in terms of FID, compared with the unconditional face generation methods (ProgressiveGAN~\cite{Karras_2018} and StyleGAN~\cite{Karras_2019}), our generated images are still qualitatively poorer.
To further improve our method, we can adopt the progressive training approach from the aforementioned methods. 
We first train a small portion of the model to produce a good quality image in a small resolution, and then gradually add the rest of the model to produce higher resolution images.
By doing so, we can stabilize the training process to produce images with better visual quality with higher resolution.

{\small
\bibliographystyle{ieee_fullname}
\bibliography{reference}
}

\end{document}